\def\year{2019}\relax
\documentclass[letterpaper]{article} 
\usepackage{arxiv}  
\usepackage{times}  
\usepackage{helvet}  
\usepackage{courier}  
\usepackage{url}  
\usepackage{graphicx}  

\usepackage{amsmath,amssymb} 
\usepackage{color}
\usepackage{framed}
\usepackage{enumitem}
\usepackage{ctable}
\usepackage{subfigure}
\usepackage{multirow}
\usepackage{xspace}

\usepackage{flushend}
 
\makeatletter
\def\onedot{\futurelet\@let@token\@onedot}
\def\@onedot{\ifx\@let@token.\else.\null\fi\xspace}

\def\eg{\emph{e.g}\onedot} 
\def\ie{\emph{i.e}\onedot} 
 
\def\etc{\emph{etc}\onedot}

\makeatother

\newcommand{\bR}{{\bf R}}
\newcommand{\bt}{{\bf t}}
\newcommand{\bX}{{\bf X}}

\newcommand{\bp}{{\bf p}}
\newcommand{\rT}{\mathrm{T}}
\newcommand{\cV}{\mathcal{V}}
\newcommand{\cS}{\mathcal{S}}
\newcommand{\cH}{\mathcal{H}}

\begin{document}
%
\title{Deep Single-View 3D Object Reconstruction with Visual Hull Embedding}
\author{Hanqing Wang\thanks{Part of this work was done when HW was an intern at MSR.}$^{1}$ \quad Jiaolong Yang$^{2}$ \quad Wei Liang$^{1}$ \quad Xin Tong$^{2}$ \\
$^1${Beijing Institute of Technology} \quad  $^2${Microsoft Research} \\
{\tt\small \{hanqingwang,liangwei\}@bit.edu.cn \{jiaoyan,xtong\}@microsoft.com}}

\maketitle

\begin{abstract}
3D object reconstruction is a fundamental task of many robotics and AI problems. With the aid of deep convolutional neural networks (CNNs), 3D object reconstruction has witnessed a significant progress in recent years. However, possibly due to the prohibitively high dimension of the 3D object space, the results from deep CNNs are often prone to missing some shape details. In this paper, we present an approach which aims to preserve more shape details and improve the reconstruction quality. The key idea of our method is to leverage object mask and pose estimation from CNNs to assist the 3D shape learning by constructing a probabilistic single-view visual hull inside of the network. Our method works by first predicting a coarse shape as well as the object pose and silhouette using CNNs, followed by a novel 3D refinement CNN which refines the coarse shapes using the constructed probabilistic visual hulls. Experiment on both synthetic data and real images show that embedding a single-view visual hull for shape refinement can significantly improve the reconstruction quality by recovering more shapes details and improving shape consistency with the input image.
\end{abstract}

\section{Introduction}
Recovering the dense 3D shapes of objects from 2D imageries is a fundamental AI problem which has many applications such as robot-environment interaction, 3D-based object retrieval and recognition, \etc. Given a single image of an object, a human can reason the 3D structure of the object reliably. However, single-view 3D object reconstruction is very challenging for computer algorithms. 

Recently, a significant progress of single-view 3D reconstruction has been achieved by using deep convolutional neural networks (CNNs) \cite{choy20163d,girdhar2016learning,wu2016learning,yan2016perspective,fan2017point,tulsiani2017multi,zhu2017rethinking,wu2017marrnet,tulsiani2018multi}. Most CNN-based methods reconstruct the object shapes using 2D and 3D convolutions in a 2D encoder-3D decoder structure with the volumetric 3D representation. The input to these CNNs are object images taken under unknown viewpoints, while the output shapes are often aligned with the canonical viewpoint in a single, pre-defined 3D coordinate system such that shape regression is more tractable.
 
\begin{figure}[t!]
	\centering
	\includegraphics[width=\columnwidth]{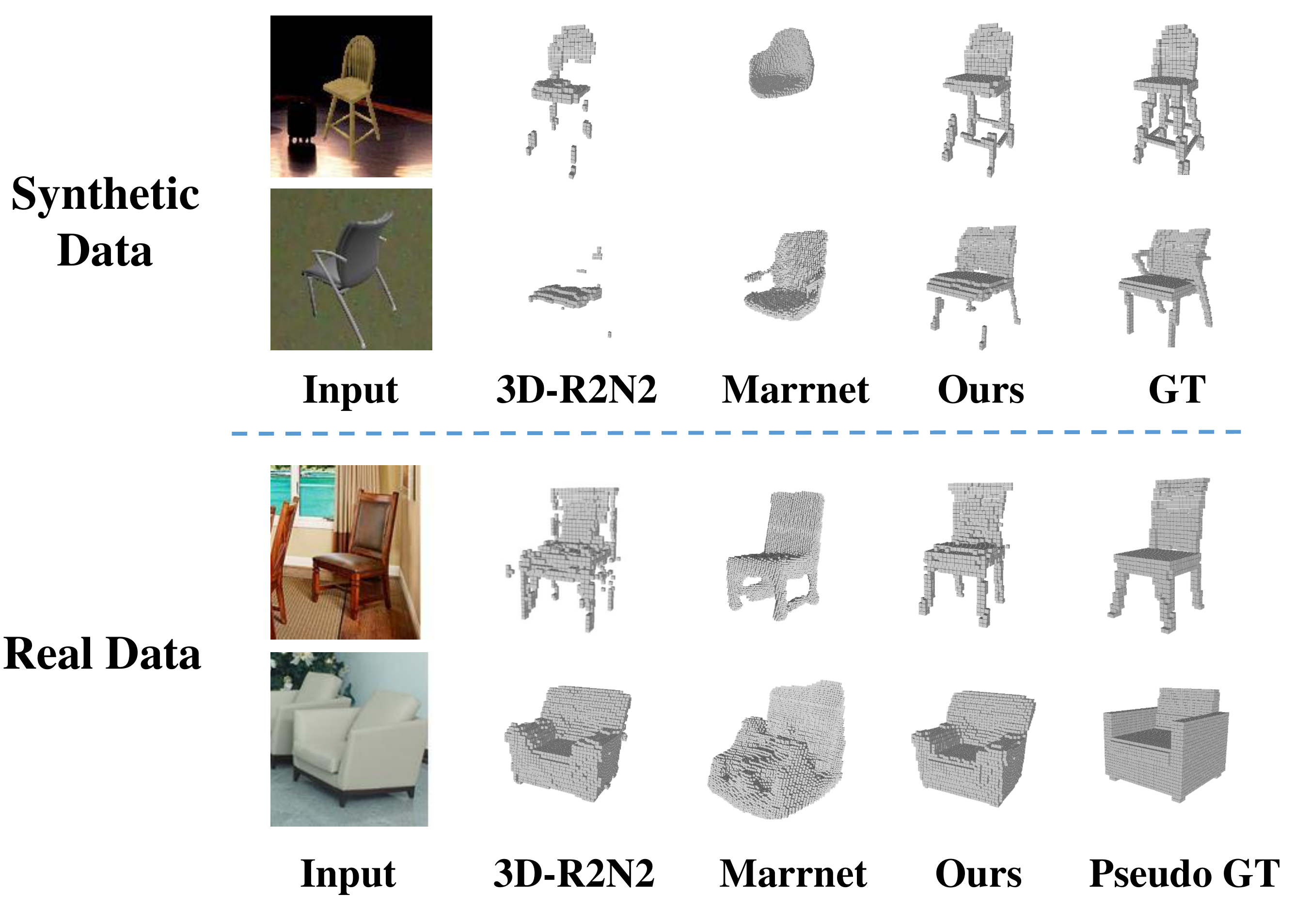}
	\vspace{-18pt}
	\caption{Some results reconstructed from synthetic data and real data by the baseline approach~\cite{choy20163d}, MarrNet~\cite{wu2017marrnet} and our approach on the chair category. Note the inconsistency with input images and missing parts in the results of the former two methods.\label{fig:teaser}\vspace{-18pt}}
\end{figure}

Although promising results have been shown by these CNN-based methods, single-view 3D reconstruction is still a challenging problem and the results are far from being perfect. One of the main difficulties lies in the object shape variations which can be very large even in a same object category. The appearance variations in the input images caused by pose differences make this task even harder. Consequently, the results from CNN-based methods are prone to missing some shape details and sometimes generate plausible shapes which, however, are inconsistent with input images, as shown in Figure~\ref{fig:teaser}.
 
In this paper, we propose an approach to improve the fidelity of the reconstructed shapes by CNNs. Our method combined traditional wisdom into the network architecture. It is motivated by two observations: 1) while directly recovering all the shape details in 3D is difficult, extracting the projected shape silhouette on the 2D plane, \ie segmenting out the object from background in a relatively easy task using CNNs; 2) for some common objects such as chairs and cars whose 3D coordinate systems are well defined without ambiguity, the object pose (or equivalently, the viewpoint) can also be well estimated by a CNN~\cite{su2015render,massa2016crafting}. As such, we propose to leverage the object silhouettes to assist the 3D learning by lifting them to 3D using pose estimates.

\begin{figure}[t!]
	\centering
	\includegraphics[width=0.9\columnwidth]{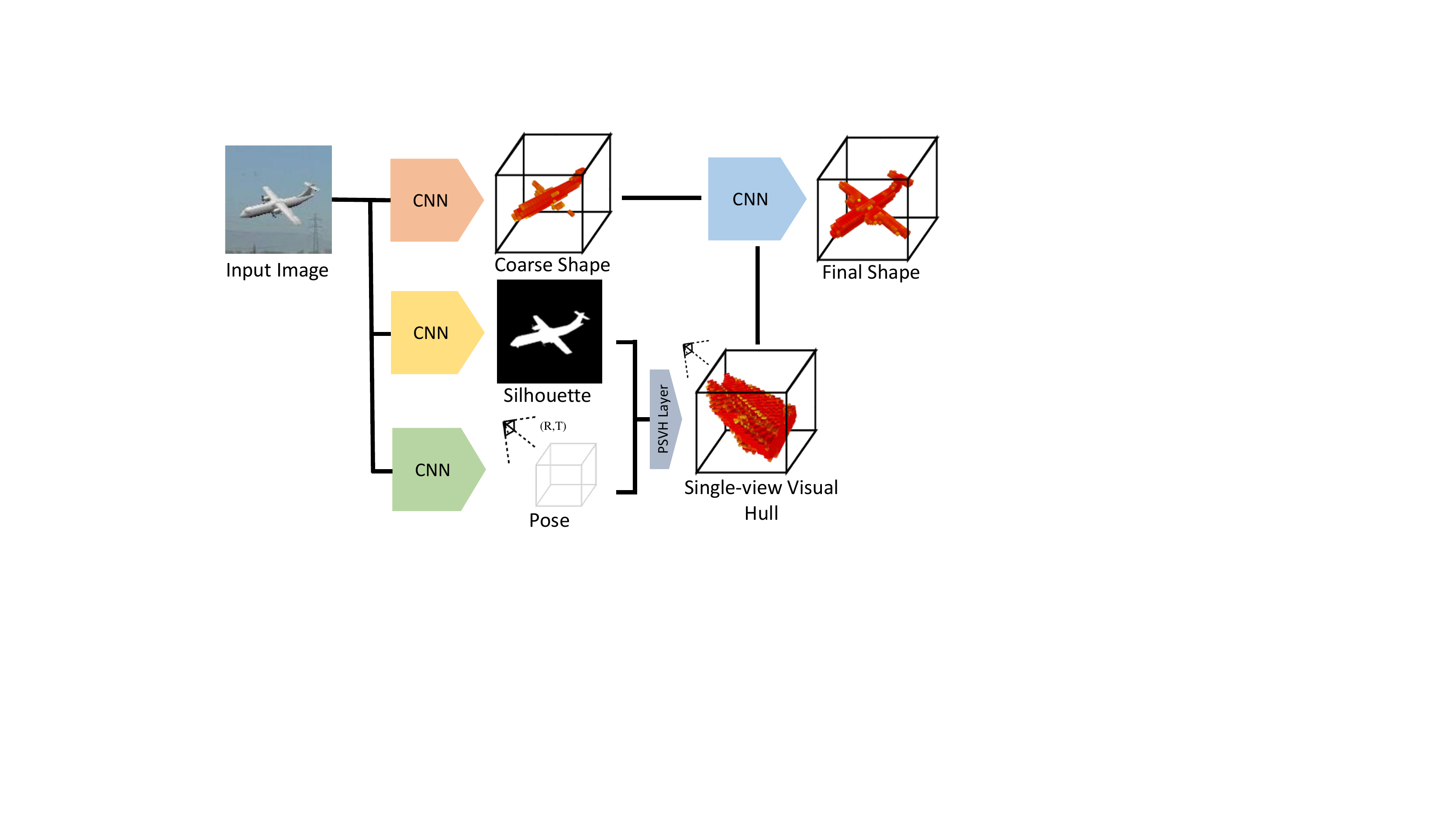}
	\vspace{-8pt}
	\caption{An overview of the proposed method. Given an input image, we first use CNNs to predict a coarse 3D volumetric shape, the silhouette and the object pose. The latter two are used to construct a single-view visual hull, which is used to refine the coarse shape using another CNN.\label{fig:framework}\vspace{-15pt}}
\end{figure}

Figure~\ref{fig:framework} is a schematic description of our method, which is a pure GPU-friendly neural network solution. Specifically, we embed into the network a single-view visual hull using the estimated object silhouettes and poses. Embedding a visual hull can help recover more shape details by considering the projection relationship between the reconstructed 3D shape and the 2D silhouette. Since both the pose and segmentation are subject to estimation error, we opted for a ``soft'' visual-hull embedding strategy: we first predict a coarse 3D shape using a CNN, then employ another CNN to refine the coarse shape with the constructed visual hull. We propose a  probabilistic single-view visual hull (PSVH) construction layer which is differentiable such that the whole network can be trained end-to-end.

In summary, we present a novel CNN-based approach which uses a single-view visual hull to improve the quality of shape predictions. Through our method, the perspective geometry is seamlessly embedded into a deep network. We evaluate our method on synthetic data and real images, and demonstrate that using a single-view visual hull can significantly improve the reconstruction quality by recovering more shape details and improving shape consistency with input images.

\section{Related Work}


\vspace{6pt}
\noindent\emph{Traditional methods.~} Reconstructing a dense 3D object shape from a single image is an ill-posed problem. Traditional methods resort to geometry priors for the otherwise prohibitively challenging task. For example, some methods leveraged pre-defined CAD models \cite{sun2013object}. Category-specific reconstruction methods \cite{vicente2014reconstructing,kar2015category,tulsiani2017learning} reconstruct a 3D shape template from images of the objects in the same category as shape prior. Given an input image, these methods estimate silhouette and viewpoint from the input image and then reconstruct 3D object shape by fitting the shape template to the estimated visual hull. Our method integrates the single-view visual hull with deep neural network for reconstructing 3D shape from single image. 

\vspace{6pt}
\noindent\emph{Deep learning for 3D reconstruction.~} Deep learning based methods directly learn the mapping from 2D image to a dense 3D shape from training data.
For example, \cite{choy20163d} directly trained a network with 3D shape loss. \cite{yan2016perspective} trained a network by minimizing the difference between the silhouette of the predicted 3D shape and ground truth silhouette on multiple views. A ray consistency loss is proposed in \cite{tulsiani2017multi} which uses other types of multi-view observations for training such as depth, color and semantics. \cite{wu2017marrnet} applied CNNs to first predict the 2.5D sketches including normal, depth and silhouette, then reconstruct the 3D shape. A reprojection consistency constraint between the 3D shape and 2.5D sketches is used to finetune the network on real images. \cite{zhu2017rethinking} jointly trained a pose regressor with a 3D reconstruction network so that the object images with annotated masks yet unknown poses can be used for training.
Many existing methods have explored using pose and silhouette (or other 2D/2.5D observations) to supervise the 3D shape prediction~\cite{yan2016perspective,tulsiani2017multi,gwak2017weakly,zhu2017rethinking,wu2017marrnet,tulsiani2018multi}. However, our goal is to refine the 3D shape inside of the network using an estimated visual hull, and our visual hull construction is an inverse process of their shape-to-image projection scheme. More discussions can be found in the \emph{supplementary material}.
 
\vspace{6pt}
\noindent\emph{Generative models for 3D shape.~} Some efforts are devoted to modeling the 3D shape space using generative models such as GAN~\cite{goodfellow2014generative} and VAE~\cite{kingmaw13vae}. In \cite{wu2016learning}, a 3D-GAN method is proposed for learning the latent space of 3D shapes and a 3D-VAE-GAN is also presented for mapping image space to shape space. 
A fully convolutional 3D autoencoder for learning shape representation from noisy data is proposed in \cite{sharma2016vconv}.
A weakly-supervised GAN for 3D reconstruction with the weak supervision from silhouettes can be found in \cite{gwak2017weakly}.

\vspace{6pt}
\noindent\emph{3D shape representation.} Most deep object reconstruction methods use the voxel grid representation~\cite{choy20163d,girdhar2016learning,yan2016perspective,tulsiani2017multi,zhu2017rethinking,wu2017marrnet,wu2016learning,gwak2017weakly}, \ie, the output is a voxelized occupancy map. Recently, memory-efficient representations such as point clouds~\cite{qi2017pointnet}, voxel octree~\cite{hane2017hierarchical,tatarchenko2017octree} and shape primitives~\cite{zou20173d} are investigated.


\vspace{6pt}
\noindent\emph{Visual hull for deep multi-view 3D reconstruction.~} Some recent works use visual hulls of color~\cite{ji2017surfacenet} or learned feature~\cite{kar2017learning} for multi-view stereo with CNNs. Our method is different from theirs in several ways. First, the motivations of using visual hulls differ: they use visual hulls as input to their multi-view stereo matching networks in order to reconstruct the object shape, 
whereas our goal is to leverage a visual hull to refine a coarse single-view shape prediction. Second, the object poses are given in their methods, while in ours the object pose is estimated by a CNN. Related to the above, our novel visual hull construction layer is made differentiable, and object segmentation, pose estimation and 3D reconstruction are jointly trained in one framework.

\section{Our Approach}

In this section, we detail our method which takes as input a single image of a common object such as car, chair and coach, and predicts its 3D shape. We assume the objects are roughly centered (\eg those in bounding boxes given by an object detector).

\vspace{7pt}
\noindent\emph{Shape representation.~} We use voxel grid for shape representation similar to previous works~\cite{wu20153d,yan2016perspective,wu2016learning,zhu2017rethinking,wu2017marrnet}, \ie, the output of our network is a voxelized occupancy map in the 3D space. This representation is very suitable for visual hull construction and processing, and it is also possible to extend our method to use tree-structured voxel grids for more fine-grained details~\cite{hane2017hierarchical,tatarchenko2017octree}

\vspace{7pt}
\noindent\emph{Camera model.~}\label{sec:cameramodel}
We choose the perspective camera model for the 3D-2D projection geometry, and reconstruct the object in a unit-length cube located in front of the camera (\ie, with cube center near the positive Z-axis in the camera coordinate frame). Under a perspective camera model, the relationship between a 3D point $(X,Y,Z)$ and its projected pixel location $(u,v)$ on the image is
\begin{equation}\label{eq:projection}
	Z[u,v,1]^T = \mathbf{K}\big(\bR[X,Y,Z]^\rT + \bt\big)
\end{equation}
where $\mathbf{K} =\scriptsize \left[ {\begin{array}{ccc}
	f&0&u_0 \\
	0&f&v_0 \\
	0&0&1 \\
	\end{array} } \right]$ is the camera intrinsic matrix with $f$ being the focal length and $(u_0,v_0)$ the principle point. 
We assume that the principal points coincide with image center (or otherwise given), and focal lengths are known. Note that when the exact focal length is not available, a rough estimate or an approximation may still suffice. When the object is reasonably distant from the camera, one can use a large focal length to strike a balance between perspective and weak-perspective models. 

\vspace{7pt}
\noindent\emph{Pose parametrization.~} The object pose is characterized by a rotation matrix $\bR\in \mathrm{SO(3)}$ and a translation vector $\bt=[t_X,t_Y,t_Z]^\rT\in\mathbb{R}^3$ in Eq.~\ref{eq:projection}. We parameterize rotation simply with Euler angles $\theta_i,i=1,2,3$. For translation we estimate $t_Z$ and a 2D vector $[t_u,t_v]$ which centralizes the object on image plane, and obtain $\bt$ via $[\frac{t_u}{f}*t_Z, \frac{t_v}{f}*t_Z, t_Z]^\rT$.
In summary, we parameterize the pose as a 6-D vector $\bp=[\theta_1,\theta_2,\theta_3,t_u,t_v, t_Z]^\rT$.

\begin{figure*}[t!]
	\centering
	\includegraphics[width=1.0\textwidth]{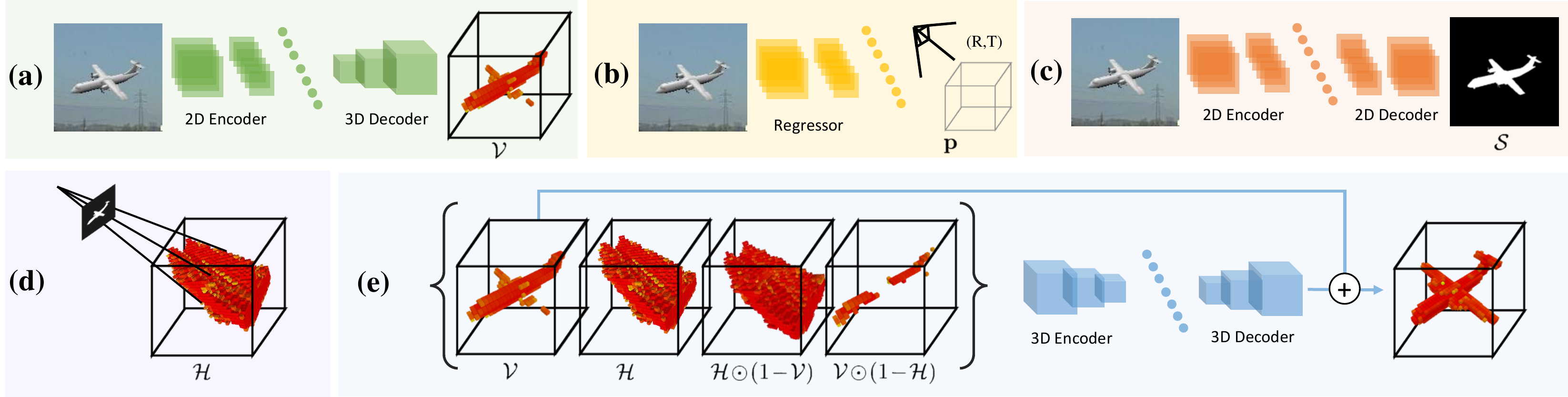}
	\vspace{-18pt}
	\caption{Network structure illustration. Our network consists of (a) a coarse shape estimation subnet V-Net, (b) an object pose estimation subnet P-Net , (c) an object segmentation subnet, (d) a probabilistic single-view visual hull (PSVH) layer, and finally (e) a shape refinement network R-Net.}\label{fig:network}
	\vspace{-12pt}
\end{figure*}

\subsection{Sub-nets for Pose, Silhouette and Coarse Shape}\label{sec:threesubnets}

Given a single image as input, we first apply a CNN to directly regress a 3D volumetric reconstruction similar to previous works such as \cite{choy20163d}. We call this network the \emph{V-Net}. Additionally, we apply another two CNNs for pose estimation and segmentation, referred to as \emph{P-Net} and \emph{S-Net} respectively. In the following we describe the main structure of these sub-networks; more details can be found in the \emph{supplementary material}. 
 
\vspace{3pt}
\emph{V-Net:} The V-Net for voxel occupancy prediction consists of a 2D encoder and a 3D decoder, as depicted in Fig.~\ref{fig:network}~(a). It is adapted from the network structure of \cite{choy20163d}, and the main difference is we replaced their LSTM layer designed for multi-view reconstruction with a simple fully connected (FC) layer.
We denote the 3D voxel occupation probability map produced by the V-Net as $\cV$.
 
\vspace{3pt}
\emph{P-Net:} The P-Net for pose estimation is a simple regressor outputting 6-dimensional pose vector denoted as $\bp$, as shown in Fig.~\ref{fig:network}~(b). We construct the P-Net structure simply by appending two FC layers to the encoder structure of V-Net, one with 512 neurons and the other with 6 neurons. 
 
\vspace{3pt}
\emph{S-Net:} The S-Net for object segmentation has a 2D encoder-decoder structure, as shown in Fig.~\ref{fig:network}~(c). We use the same encoder structure of V-Net for S-Net encoder, and apply a mirrored decoder structure consisting of deconv and uppooling layers. The S-Net generates an object probability map of 2D pixels, which we denote as $\cS$.

\subsection{PSVH Layer for Probabilistic Visual Hull}

Given the estimated pose $\bp$ and the object probability map $\cS$ on the image plane, we construct inside of our neural network a Probabilistic Single-view Visual Hull (PSVH) in the 3D voxel grid. 
To achieve this, we project each voxel location $\mathbf{X}$ onto the image plane by the perspective transformation in Eq.~\ref{eq:projection} to obtain its corresponding pixel $\mathbf{x}$. Then we assign $\cH(\mathbf{X})=\cS(\mathbf{x})$, where $\cH$ denotes the generated probabilistic visual hull.
This process is illustrated in Fig.~\ref{fig:network} (d).

The PSVH layer is differentiable, which means that the gradients backpropagated to $\cH$ can be backpropagated to $\cS$ and pose $P$, hence further to S-Net and P-Net. The gradient of $\cH$ with respect to $\cS$ is easy to discern: we have built correspondences from $\cH$ to $\cS$ and simply copied the values. Propagating gradients to $\bp$ is somewhat tricky. According to the chain rule, we have 
$\frac{\partial l}{\partial \bp}=\sum_\bX\frac{\partial l}{\partial\cH(\bX)}\frac{\partial \cH(\bX)}{\partial\bX}\frac{\partial \bX}{\partial \bp}$
where $l$ is the network loss. Obtaining $\frac{\partial l}{\partial \bp}$ necessitates computing $\frac{\partial \cH(\bX)}{\partial\bX}$, \ie, the spatial gradients of $\cH$, which can be numerically computed by three convolution operations with pre-defined kernels along X-, Y- and Z-axis respectively. $\frac{\partial \bX}{\partial \bp}$ can be derived analytically.


\subsection{Refinement Network for Final Shape}

With a coarse voxel occupancy probability $\cV$ from V-Net and the visual hull $\cH$ from the PSVH layer, we use a 3D CNN to refine $\cV$ and obtain a final prediction, denoted by $\cV^{+}$. We refer to this refinement CNN as \emph{R-Net}. The basic structure of our R-Net is shown in Fig.~\ref{fig:network} (e). It consists of five 3D conv layers in the encoder and 14 3D conv layers in the decoder. 

A straightforward way for R-Net to process $\cV$ and $\cH$ is concatenating $\cV$ and $\cH$ to form a 2-channel 3D voxel grid as input then generating a new $\cV$ as output. Nevertheless, we have some domain knowledge on this specific problem. For example, if a voxel predicted as occupied falls out of the visual hull, it's likely to be a false alarm; if the prediction does not have any occupied voxel in a viewing ray of the visual hull, some voxels may have been missed. This domain knowledge prompted us to design the R-Net in the following manners.

First, in addition to $\cV$ and $\cH$, we feed into R-Net two more occupancy probability maps: $\cV\odot(1-\cH)$ and $\cH\odot(1-\cV)$ where $\odot$ denotes element-wise product. These two probability maps characterize voxels in $\cV$ but not in $\cH$, and voxels in $\cH$ but not in $\cV$\,\footnote{A better alternative for $\cH\odot(1-\cV)$ would be constructing another visual hull using $\cV$ and $\bp$ then compute its difference from $\cH$. We choose $\cH\odot(1-\cV)$ here for simplicity.}, respectively. 
Second, we add a residual connection between the input voxel prediction $\cV$ and the output of the last layer. This way, we guide R-Net to generate an effective \emph{shape deformation} to refine $\cV$ rather than directly predicting a new $\cV$, as the predicted $\cV$ from V-Net is often mostly reasonable (as found in our experiments).


\subsection{Network Training}\label{sec:training}

We now present our training strategies, including the training pipeline for the sub-networks and their training losses.

\vspace{7pt}
\noindent\emph{Training pipeline.~} We employ a three-step network training algorithm to train the proposed network. Specifically, we first train V-Net, S-Net and R-Net separately, with input training images and their ground-truth shapes, silhouettes and poses. After V-Net converges, we train R-Net independently, with the predicted voxel occupancy probability $\cV$ from V-Net and the ground-truth visual hull, which is constructed by ground-truth silhouettes and poses via the PSVH layer. The goal is to let R-Net learn how to refine coarse shape predictions with ideal, error-free visual hulls. In the last stage, we finetune the whole network, granting the subnets more opportunity to cooperate accordingly. Notably, the R-Net will adapt to input visual hulls that are subject to estimation error from S-Net and P-Net.

\vspace{7pt}
\noindent\emph{Training loss.~}
We use the binary cross-entropy loss to train V-Net, S-Net and R-Net. Concretely, let $p_n$ be the estimated probability at location $n$ in either $\cV$, $\cS$ or $\cV^+$, then the loss is defined as
\begin{equation}\label{eq:1}
	l=-\frac{1}{N}\sum_{n} \big(p_n^*\log p_n + (1-p_n^*)log(1-p_n)\big)
\end{equation}
where $p_n^*$ is the target probability (0 or 1). $n$ traverses over the 3D voxels for V-Net and R-Net, and over 2D pixels for S-Net. The P-Net produces a 6-D pose estimate $\bp=[\theta_1,\theta_2,\theta_3,t_u,t_v, t_Z]^\rT$ as described before. We use the $L_1$ regression loss to train the network:
\begin{equation}\label{eq:2}
	l=\sum_{i=1,2,3}\!\!\!\alpha|\theta_i\!-\!\theta_i^{*}|\, + \sum_{j=u,v}\!\!\beta|t_j\!-\!t_j^{*}|\, +\, \gamma|t_Z\!-\!t_Z^{*}|,
\end{equation}
where the Euler angles are normalized into $[0,1]$. We found in our experiments the $L_1$ loss produces better results than an $L_2$ loss.



\section{Experiments}


\vspace{7pt}
\noindent\emph{Implementation details.~}
Our network is implemented in TensorFlow. The input image size is $128\times128$ and the output voxel grid size is $32\!\times\!32\!\times\!32$. Batch size of 24 and the ADAM solver are used throughout the training. We use a learning rate of $1e\!-\!4$ for S-Net, V-Net and R-Net and divide it by 10 at the 20K-th and 60K-th iterations. The learning rate for P-Net is $1e\!-\!5$ and is dropped by 10 at the 60K-th iteration. When finetuning all the subnets together the learning rate is $1e\!-\!5$ and dropped by 10 at the 20K-th iteration.

\vspace{7pt}
\noindent\emph{Training and testing data.~} In this paper, we test our method on four common object categories: \emph{car} and \emph{airplane} as the representative vehicle objects, and \emph{chair} and \emph{couch} as furniture classes. Real images that come with precise 3D shapes are difficult to obtain, so we first resort to the CAD models from the ShapeNet repository~\cite{shapenet2015}. We use the ShapeNet object images rendered by \cite{choy20163d} to train and test our method. We then use the PASCAL 3D+ dataset~\cite{xiang2014beyond} to evaluate our method on real images with pseudo ground truth shapes.



\begin{figure*}[t!]
	\centering
	\vspace{-5pt}
	\includegraphics[width=0.85\textwidth]{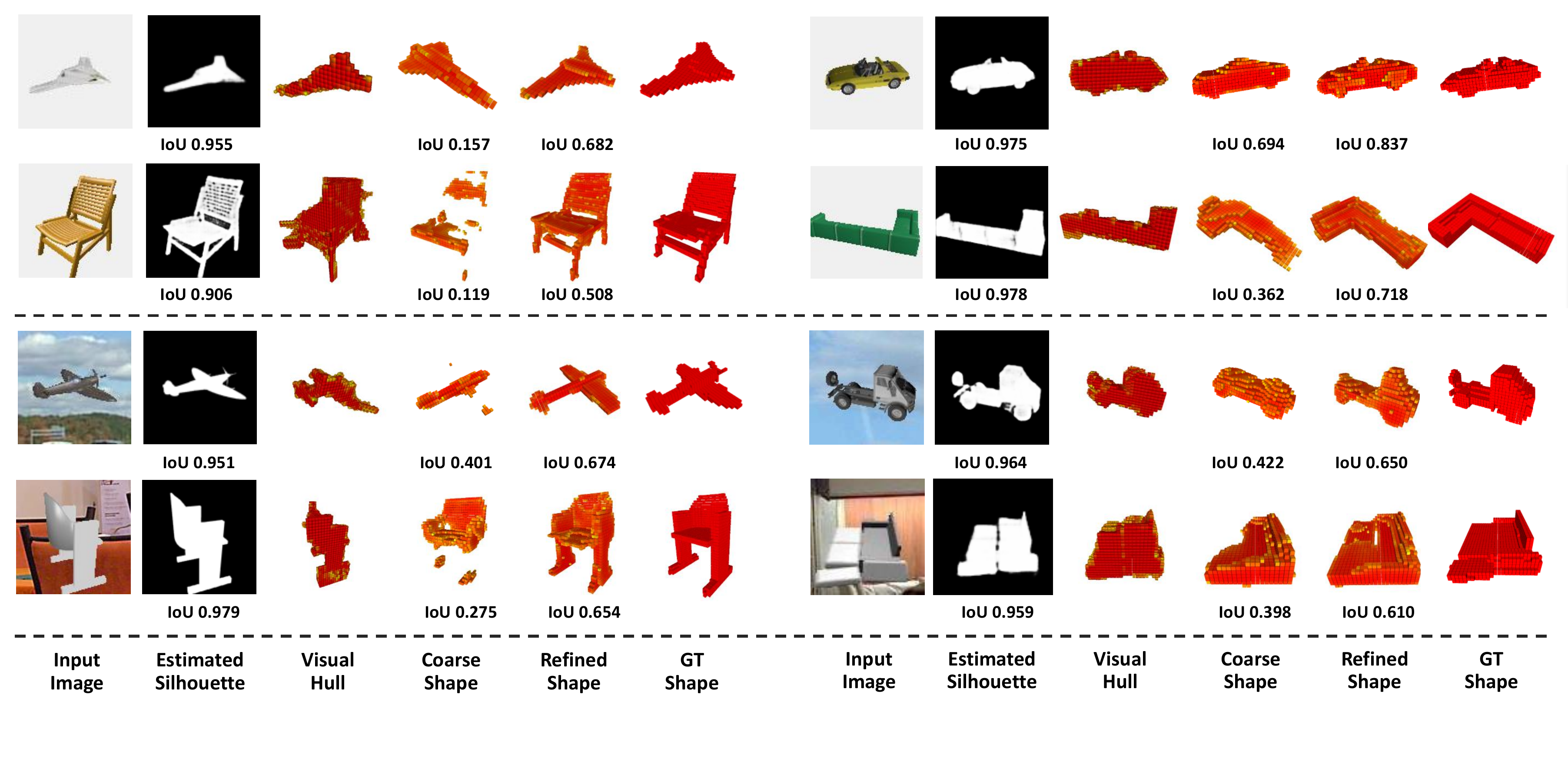}
	\vspace{-30pt}
	\caption{Qualitative results on the test set of the rendered ShapeNet objects. The top two rows and bottom two rows show some results on images with a clean background and real-image background, respectively. The color of the voxels indicates the predicted probability (red/yellow for high/low probability). While the coarse shapes may miss some shape details or be inconsistent with the image, our final shapes refined with single-view visual hulls are of high fidelity. }\label{fig:visual_results}
	\vspace{-17pt}
\end{figure*}

\subsection{Results on Rendered ShapeNet Objects}
The numbers of 3D models for the four categories are 7,496 for car, 4,045 for airplane, 6,778 for chair and 3,173 for table, respectively. In the rendering process of \cite{choy20163d}, the objects were normalized to fit in a radius-0.5 sphere, rotated with random azimuth and elevation angles, and placed in front of a 50-degree FOV camera.  Each object has 24 images rendered with random poses and lighting.

Following~\cite{choy20163d}, we use 80\% of the 3D models for training and the rest 20\% for testing. We train one network for all the four shape categories until the network converge. The rendered images are with clean background (uniform colors). During training, we blend half of the training images with random crops of natural images from the SUN database~\cite{xiao2010sun}. We binarize the output voxel probability with threshold $0.4$ and report Intersection-over-Union (IoU).


\begin{table}[!t]
	\setlength{\tabcolsep}{1.3mm}
	\centering
	\small
	\caption{The performance (shape IoU) of our method on the test set of the rendered ShapeNet objects. $\cH$ denotes visual hull and GT indicates ground truth.}\label{tab:results_2dr2n2}
	\begin{tabular}{lccccc}
		\specialrule{.1em}{0.1em}{0.1em}
		& car & \!\!\!airplane\!\!\! & chair & couch & Mean\\
		\specialrule{.1em}{0.1em}{0.1em}
		Before Refine. & 0.819 & 0.537 & 0.499 & 0.667 & 0.631 \\
		After Refine. & \textbf{0.839} & \textbf{0.631} & \textbf{0.552} & 0.698 & \textbf{0.680} \\
		
		\specialrule{.03em}{0.1em}{0.1em}
		Refine. w/o $\cH$  & 0.824 & 0.541 & 0.505 & 0.675 & 0.636 \\
		Refine. w. GT $\cH$~ & 0.869 & 0.701 & 0.592 & 0.741 & 0.726  \\
		Refine. w/o 2 prob.maps & 0.840 & 0.610 & 0.549 & 0.701 & 0.675 \\
		Refine. w/o end-to-end & 0.822 & 0.593 & 0.542 & 0.677 & 0.658 \\
		\specialrule{.1em}{0.1em}{0.1em}
	\end{tabular}
	\vspace{-14pt}
\end{table}

\begin{table}[!t]
	\setlength{\tabcolsep}{1.3mm}
	\centering
	\small
	\caption{The performance (shape IoU) of our method and PointOutNet~\cite{fan2017point}.}\label{tab:results_2dr2n2_compare}
	\begin{tabular}{lccccc}
		\specialrule{.1em}{0.1em}{0.1em}
		& car & \!\!\!airplane\!\!\! & chair & couch & Mean\\
		\specialrule{.1em}{0.1em}{0.1em}
		\!\cite{fan2017point}\!\!\! & 0.831 & 0.601 & 0.544 & \textbf{0.708} & 0.671 \\
		Ours & \textbf{0.839} & \textbf{0.631} & \textbf{0.552} & 0.698 & \textbf{0.680} \\
		\specialrule{.1em}{0.1em}{0.1em}
	\end{tabular}
	\vspace{-14pt}
\end{table}

\begin{table}[!t]
	\vspace{-1pt}
	\centering
	\small
	\caption{The performance (shape IoU) of our method on the test set of the rendered ShapeNet objects with clean background and background from natural image crops.}\label{tab:result_background}
	\vspace{2pt}
	\begin{tabular}{lccccc}
		\specialrule{.1em}{0.1em}{0.1em}
		Background & car & airplane & chair & couch & Mean\\
		\specialrule{.1em}{0.1em}{0.1em}
		Clean & 0.839 & 0.631 & 0.552 & 0.698 & 0.680 \\
		Image crop. & 0.837 & 0.617 & 0.541 & 0.700 & 0.674 \\
		\specialrule{.1em}{0.1em}{0.1em}
	\end{tabular}
	\vspace{-14pt}
\end{table}

\begin{table}[!t]
	\vspace{-1pt}
	\setlength{\tabcolsep}{1.6mm}
	\centering
	\small
	\caption{The pose estimation and segmentation quality of our P-Net and S-Net on the rendered ShapeNet objects. Mean values are shown for each category.}\label{tab:pose_silhouette}
	\begin{tabular}{lccccc}
		\specialrule{.1em}{0.1em}{0.1em}
		& car & airplane & chair & couch & Mean\\
		\specialrule{.1em}{0.1em}{0.1em}
		Rotation error & 7.96$^\circ$ & 4.72$^\circ$ & 6.59$^\circ$ & 10.41$^\circ$ & 7.42$^\circ$ \\
		Translation error~ & 3.33\% & 2.60\% & 3.26\% & 3.41\% & 3.15\% \\
		Silhouette IoU & 0.923 & 0.978 & 0.954 & 0.982 & 0.959 \\
		\specialrule{.1em}{0.1em}{0.1em}
	\end{tabular}
	\vspace{-14pt}
\end{table}

\begin{figure*}[t!]
	\centering
	\includegraphics[width=0.257\textwidth]{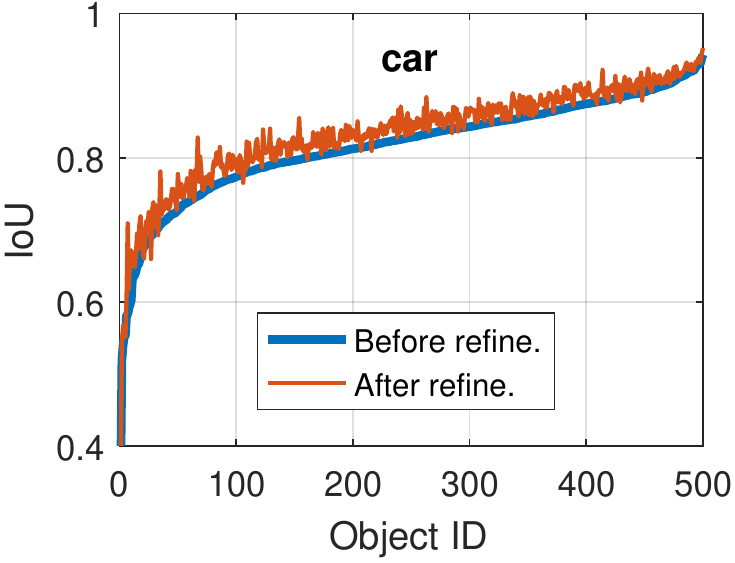}
	\includegraphics[width=0.24\textwidth]{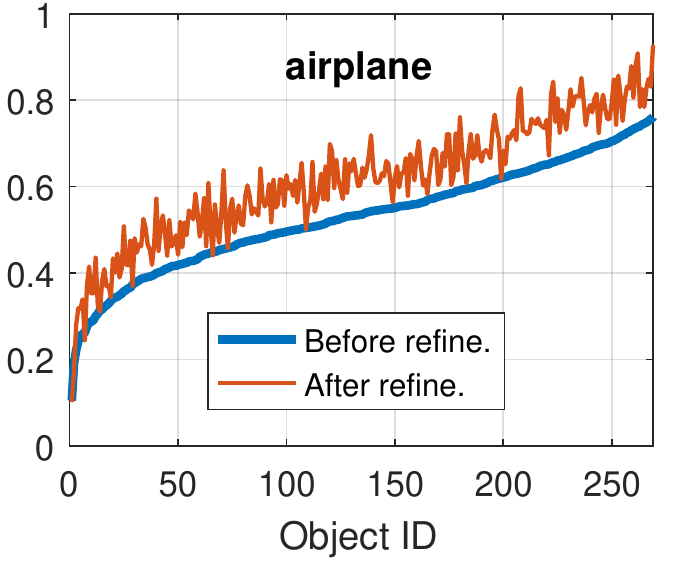}
	\includegraphics[width=0.24\textwidth]{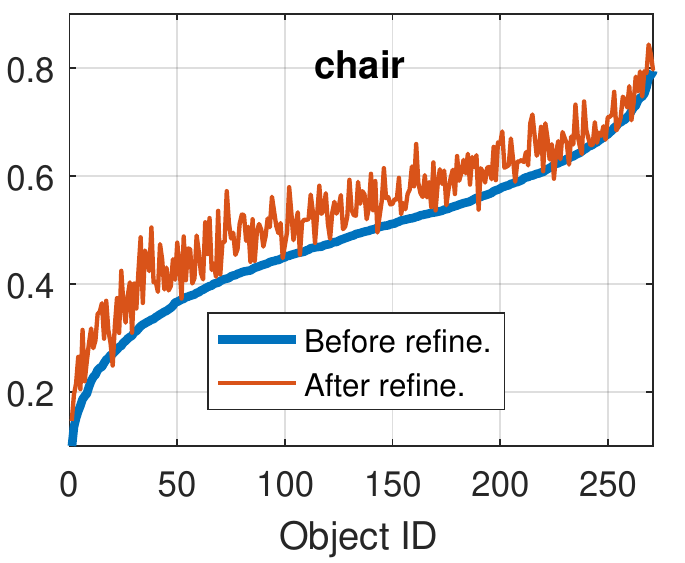}
	\includegraphics[width=0.24\textwidth]{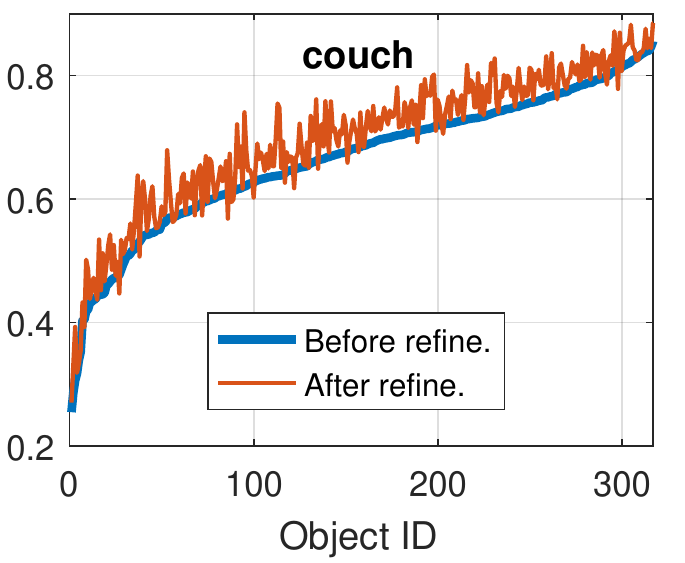}
	\vspace{-10pt}
	\caption{Comparison of the results before and after refinement on rendered ShapeNet objects.}\label{fig:iou_compare}
	\vspace{-8pt}
\end{figure*}

\vspace{7pt}
\noindent\emph{Quantitative results.~} The performance of our method evaluated by IoU is shown in Table~\ref{tab:results_2dr2n2}. 
It shows that the results after refinement (\ie, our final results) are significantly better, especially for airplane and chair where the IoUs are improved by about 16\% and 10\%, respectively. Note that since our V-Net is adapted from \cite{choy20163d} as  mentioned previously,
the results before refinement can be viewed as the 3D-R2N2 method of \cite{choy20163d} trained by us.

To better understand the performance gain from our visual hull based refinement, we compute the IoU of the coarse and refined shapes for each object from the four categories. Figure~\ref{fig:iou_compare} presents the comparisons, where the object IDs are uniformly sampled and sorted by the IoUs of coarse shapes. The efficacy of our refinement scheme can be clearly seen. It consistently benefits the shape reconstruction for most of the objects, despite none of them is seen before.

We further compare the numerical results with PointOutNet~\cite{fan2017point} which was also evaluated on this rendered dataset and used the same training/testing lists as ours.
Table~\ref{tab:results_2dr2n2} shows that our method outperformed it on the three of the four categories (car, airplane and chair) and obtained a higher mean IoU over the four categories. Note that the results of \cite{fan2017point} were obtained by first generating point clouds using their PointOutNet, then converting them to volumetric shapes and applying another 3D CNN to refine them.


Table~\ref{tab:result_background} compares the results of our method on test images with clean background and those blended with random real images. It shows that with random real image as background the results are only slightly worse. Table~\ref{tab:pose_silhouette} shows the quality of the pose and silhouette estimated by P-Net and S-Net.

\begin{figure}[t!]
	\centering
	\includegraphics[width=1\columnwidth]{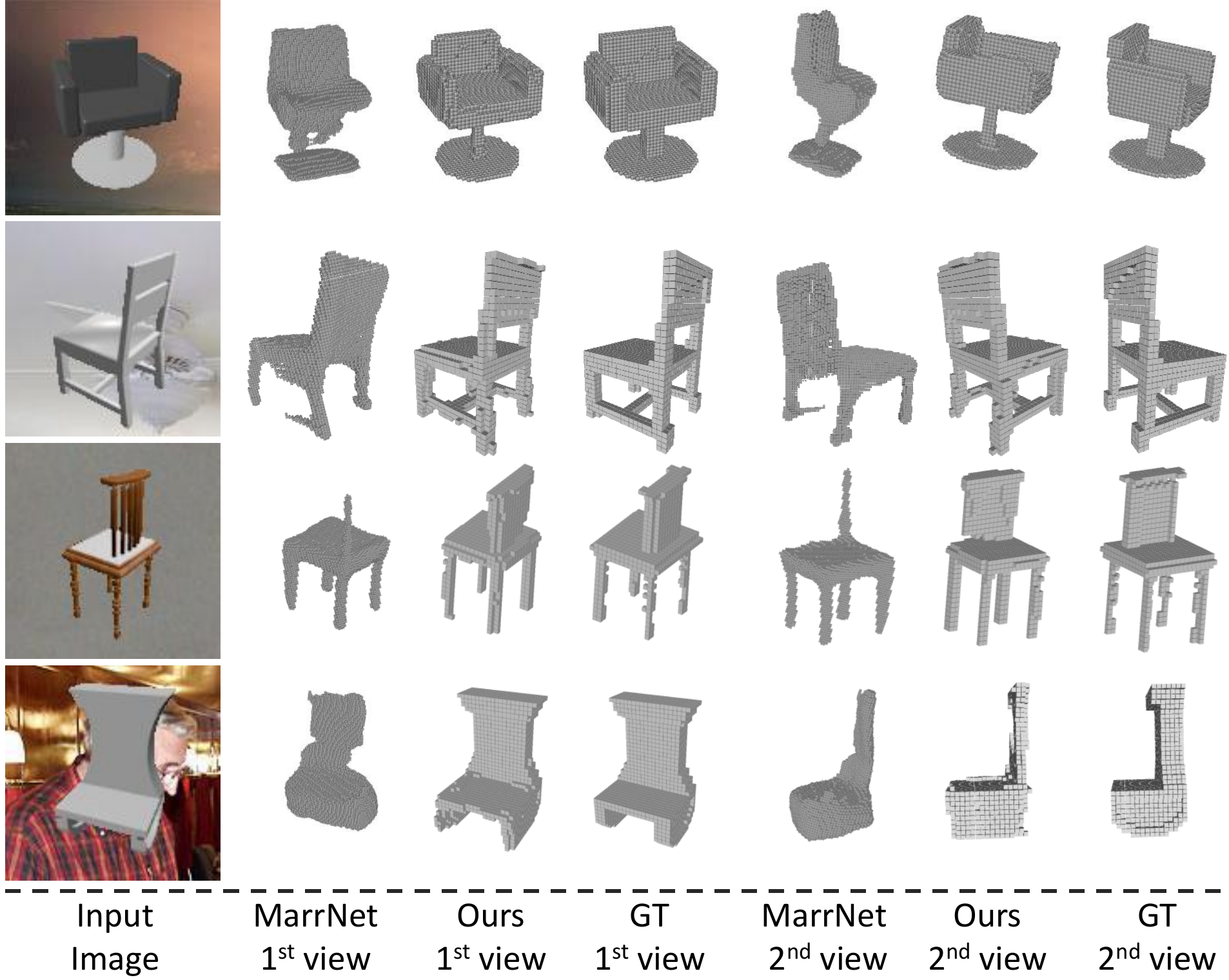}
	\vspace{-22pt}
	\caption{Result comparison with the MarrNet method on ShapeNet chairs (our testing split). Top two rows: cherry-picked results of MarrNet, compared against our results. Bottom two rows: cherry-picked results of our method, compared against MarrNet results.}\label{fig:comp_marrnet}
	\vspace{-12pt}
\end{figure}

\vspace{7pt}
\noindent\emph{Qualitative results.~} Figure~\ref{fig:visual_results} presents some visual results from our method. It can be observed that some object components especially thin structures (\eg. the chair legs in the second and fifth rows) are missed in the coarse shapes. Moreover, we find that although some coarse shapes appear quite realistic (\eg. the airplanes in the left column), they are clearly inconsistent with the input images. By leveraging the single-view visual hull for refinement, many shape details can be recovered in our final results, and they appear much more consistent with the input images.

We also compare our results qualitatively with MarrNet~\cite{wu2017marrnet}, another state-of-the-art single-view 3D object reconstruction method\footnote{We were not able to compare the results quantitatively: MarrNet directly predicts the shapes in the current camera view which are not aligned with GT shapes; moreover, the training and testing splits for MarrNet are not disclosed in \cite{wu2017marrnet}.}.
The authors released a MarrNet model trained solely on the chair category of the ShapeNet objects.
Figure~\ref{fig:comp_marrnet} presents the results on four chair images, where the first/last two are relatively good results from MarrNet/our method cherry-picked among 100 objects on our test set. It can be seen that in both cases, our method generated better results than MarrNet. Our predicted shapes are more complete and consistent with the input images.

\subsection{Results on the Pascal 3D+ Dataset}

\begin{table}[!t]
	\vspace{0pt}
	\centering
	\small
	\caption{The performance (shape IoU) of our method on the test set of the PASCAL 3D+ dataset.
	}\label{tab:results_pascal}
	\begin{tabular}{lccccc}
		\specialrule{.1em}{0.1em}{0.1em}
		& car & airplane & chair & couch & Mean\\
		\specialrule{.1em}{0.1em}{0.1em}
		Before Refine. & 0.625 & 0.647 & 0.341 & 0.633 & 0.552 \\
		After Refine. & \textbf{0.653} & \textbf{0.690} & \textbf{0.341} & \textbf{0.664} & \textbf{0.587} \\
		\specialrule{.1em}{0.1em}{0.1em}
	\end{tabular}
	\vspace{-12pt}
\end{table}


\begin{table}[!t]
	\vspace{0pt}
	\centering
	\small
	\caption{The pose estimation and segmentation quality of P-Net and S-Net on PASCAL 3D+. Median values are reported. Note that the pseudo ground-truth poses and silhouettes used for evaluation are noisy.\label{tab:pose_silhouette_pascal}}
	\vspace{0pt}
	\begin{tabular}{lcccc}
		\specialrule{.1em}{0.1em}{0.1em}
		& car & airplane & chair & couch \\
		\specialrule{.1em}{0.1em}{0.1em}
		\!\multirow{1}{*}{Rotation error} &39.4$^\circ$ &25.5$^\circ$ & 43.6$^\circ$ & 34.0$^\circ$ \\
		\hline
		\!\multirow{1}{*}{Translation error} & 8.6\% & 4.4\% & 12.7\% & 12.5\% \\
		\hline
		\!\multirow{1}{*}{Silhouette IoU}  & 0.757 & 0.614 & 0.457  & 0.696 \\
		\specialrule{.1em}{0.1em}{0.1em}
	\end{tabular}
	\vspace{-12pt}
\end{table}

We now evaluate our method on real images from the PASCAL 3D+ dataset~\cite{xiang2014beyond}. This dataset only have pseudo ground-truth shapes for real images, which makes it very challenging for our visual hull based refinement scheme. Moreover, the provided object poses are noisy due to the lack of accurate 3D shapes, making it difficult to train our pose network.

%

To test our method on this dataset, we finetune our network trained on ShapeNet objects on images in PASCAL 3D+. We simply set the focal length to be $2000$ for all images since no focal length is provided. With this fixed focal length, we recomputed the object distances using the image keypoint annotations and the CAD models through reprojection error minimization.
Due to space limitation, more details are deferred to the \emph{supplementary material}.


\vspace{7pt}
\noindent\emph{Quantitative results.~} The quantitative results of our method are presented in Table~\ref{tab:results_pascal} and Table~\ref{tab:pose_silhouette_pascal}. As can be seen in Table~\ref{tab:pose_silhouette_pascal}, the pose and silhouette estimation errors are much higher than the results on the ShapeNet objects. Nevertheless, Table~\ref{tab:results_pascal} shows that our visual hull based refinement scheme still largely improved the coarse shape from V-Net for the car, airplane and couch categories. Note again that our V-Net is almost identical to the network in the 3D-R2N2 method \cite{choy20163d}.
The refinement only yields marginal IoU increase for the chair category. We observed that the chair category on this dataset contains large intra-class shape variations (yet only 10 CAD shapes as pseudo ground truth) and many instances with occlusion; see the \emph{suppl. material} for more details.

\begin{figure}
	\centering
	\includegraphics[width=0.88\columnwidth]{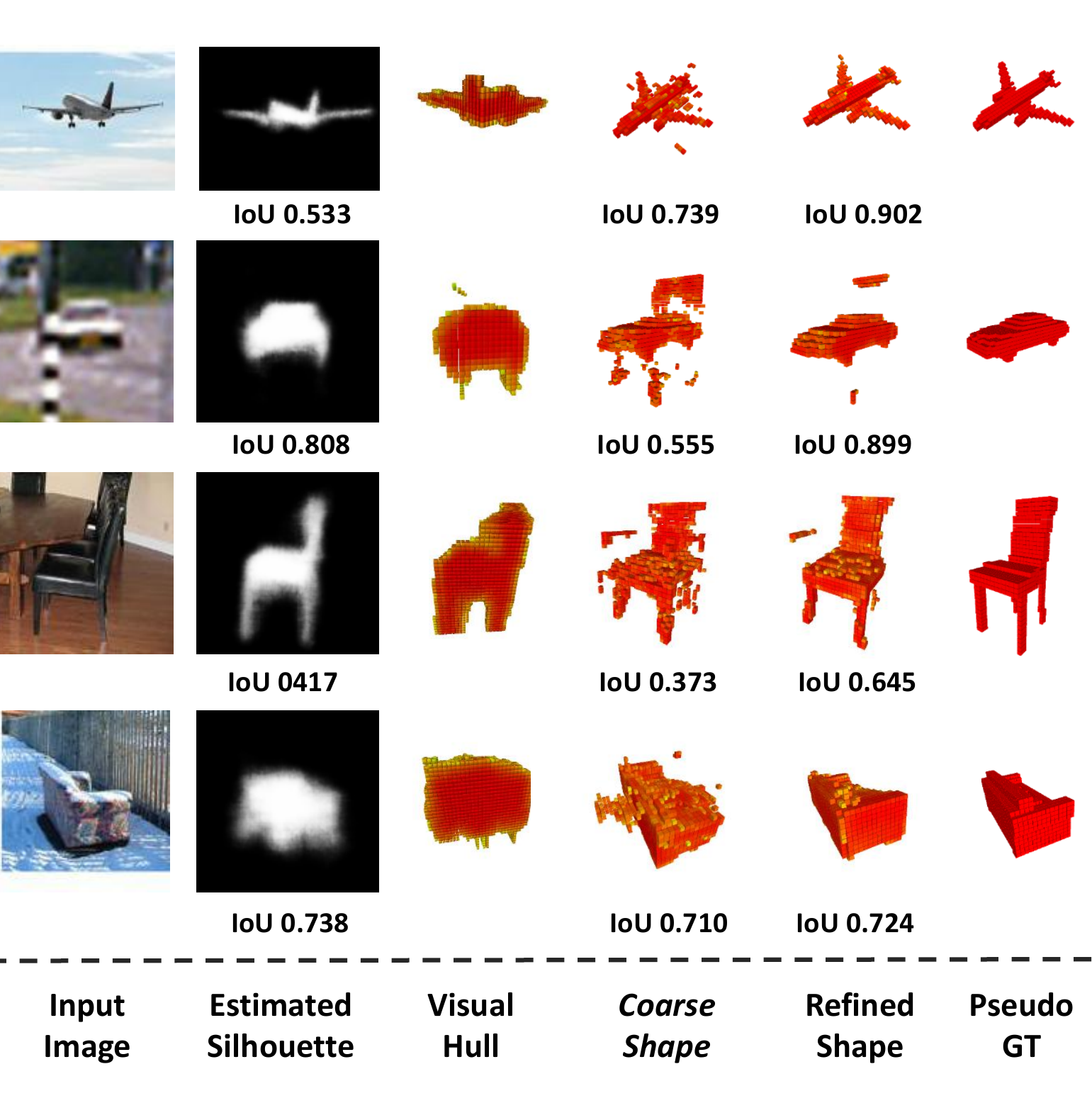}
	\vspace{-15pt}
	\caption{Qualitative results on the test set of PASCAL 3D+}\label{fig:pascal}
	\vspace{-10pt}
\end{figure}

\vspace{7pt}
\noindent\emph{Qualitative results.~}
Figure~\ref{fig:pascal} shows some visual results of our method on the test data. It can be seen that the coarse shapes are noisy or contain erroneous components. For example, possibly due to the low input image quality, the coarse shape prediction of the car image in the second row of the left column has a mixed car and chair structure. Nevertheless, the final results after the refinement are much better.

Figure~\ref{fig:teaser} shows some results from both our method and MarrNet~\cite{wu2017marrnet}. Visually inspected, our method produces better reconstruction results again.

\subsection{More Ablation Study and Performance Analysis}

\vspace{5pt}
\noindent\emph{Performance of refinement without visual hull.~} In this experiment, we remove the probabilistic visual hull and train R-Net to directly process the coarse shape. As shown in Table~\ref{tab:results_2dr2n2}, the results are slightly better than the coarse shapes, but lag far behind the results refined with visual hull.

\vspace{5pt}
\noindent\emph{Performance of refinement with GT visual hull.~} We also trained R-Net with visual hulls constructed by ground-truth poses and silhouettes. Table~\ref{tab:results_2dr2n2} shows that the performance is dramatically increased: the shape IoU is increased by up to 30\% from the coarse shape for the four object categories. The above two experiments indicate that our R-Net not only leveraged the visual hull to refine shape, but also can work remarkably well if given a quality visual hull.

\vspace{5pt}
\noindent\emph{Effect of two additional occupancy probability maps $\cV\odot(1-\cH)$ and $\cH\odot(1-\cV)$.~}  
The results in Table~\ref{tab:results_2dr2n2} shows that, if these two additional maps are removed from the input of R-Net, the mean IoU drops slightly from 0.680 to 0.675, indicating our explicit knowledge embedding helps.

\vspace{5pt}
\noindent\emph{Effect of end-to-end training.~} Table~\ref{tab:results_2dr2n2} also presents the result without end-to-end training. The clear performance drop demonstrates the necessity of our end-to-end finetuning which grants the subnets the opportunity to better adapt to each other (notably, R-Net will adapt to input visual hulls that are subject to estimation error from S-Net and P-Net).

\begin{figure}[t!]
	\centering
	\includegraphics[width=0.363\columnwidth]{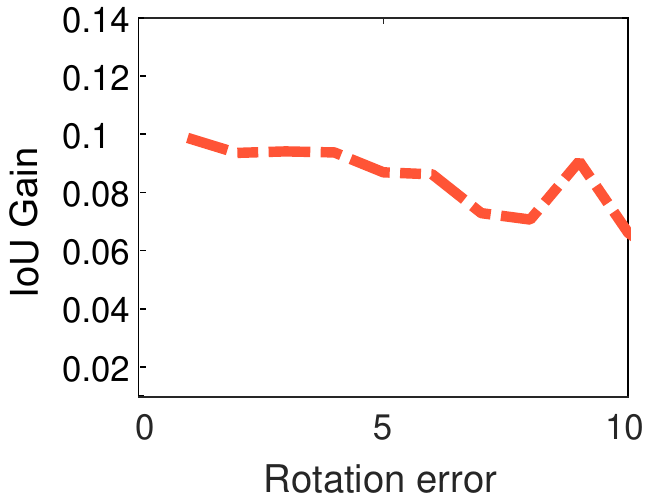}~~
	\includegraphics[width=0.278\columnwidth]{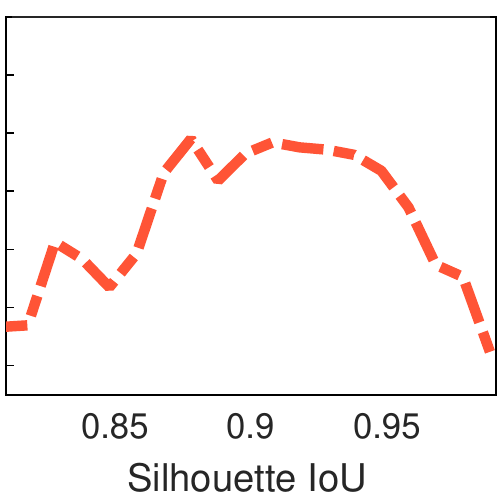}~~~
	\includegraphics[width=0.278\columnwidth]{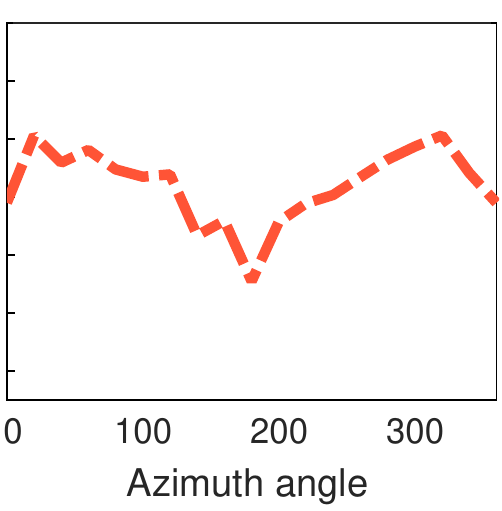}
	\vspace{-8pt}
	\caption{Performance of the shape refinement (IoU gain compared to coarse shape) \emph{w.r.t.} pose and silhouette estimation quality and rotation angle. The airplane category of the rendered ShapeNet objects is used}\label{fig:refinement_pose_silhouette}\vspace{-10pt}
\end{figure}

\vspace{5pt}
\noindent\emph{Performance w.r.t. pose and silhouette estimation quality.~}
We find that the performance gain from refinement decreases gracefully \emph{w.r.t.} rotation estimation error, as shown in Fig.~\ref{fig:refinement_pose_silhouette} (left). One interesting phenomenon is that, as shown in Fig.~\ref{fig:refinement_pose_silhouette} (middle), the best performance gain is not from best silhouette estimates. This is because larger and fuzzier silhouette estimates may compensate the 2D-to-3D correspondence errors arisen due to noisy pose estimates. 

\vspace{5pt}
\noindent\emph{Performance w.r.t. rotation angle.~} Figure~\ref{fig:refinement_pose_silhouette} (right) shows that the performance gain from refinement is high at 30 and 330 degrees while low near 0 and 180 degrees. This is easy to discern as frontal and real views exhibit more self-occlusion thus the visual hulls are less informative.


\vspace{5pt}
\noindent\emph{Sensitivity w.r.t. focal length.~} We conducted another experiment to further test our method under wrong focal lengths and distorted visual hulls. The results indicate that our method still works well with some weak-perspective approximations and the results are insensitive to the real focal lengths especially for reasonably-distant objects. The details can be found in the \emph{suppl. material}.

\subsection{Running Time}
For a batch of 24 input images, the forward pass of our whole network takes 0.44 seconds on an NVIDIA Tesla M40 GPU, \ie, our network processes one image with 18 milliseconds on average.

\section{Conclusions}

We have presented a novel framework for single-view 3D object reconstruction, where we embed the perspective geometry into a deep neural network to solve the challenging problem. Our key innovations include an in-network visual hull construction scheme which connects the 2D space and pose space to the 3D space, and a refinement 3D CNN which learns shape refinement with visual hulls. The experiments demonstrate that our method achieves very promising results on both synthetic data and real images.

\vspace{5pt}
\noindent\emph{Limitations and future work.} Since our method involves pose estimation, objects with ambiguous pose (symmetric shapes) or even do not have a well-defined pose system (irregular shapes) will be challenging. For the former cases, using a classification loss to train the pose network would be a good remedy~\cite{su2015render},
although this may render the gradient backpropagation problematic. For the latter, one possible solution is resorting to multi-view inputs and train the pose network to estimate \emph{relative} poses.

\bibliography{aaai19}
\bibliographystyle{aaai}

\end{document}